\documentclass{article}
\usepackage{PRIMEarxiv}
\usepackage{xcolor}
\usepackage[utf8]{inputenc} 
\usepackage[T1]{fontenc}    
\usepackage{hyperref}       
\usepackage{url}            
\usepackage{booktabs}       
\usepackage{amsfonts}       
\usepackage{nicefrac}       
\usepackage{microtype}      
\usepackage{lipsum}
\usepackage{fancyhdr}       
\usepackage{graphicx}       
\usepackage{amsmath} 
\usepackage{glossaries}
\graphicspath{{media/}}     

\title{An Autoencoder and Generative Adversarial Networks Approach for Multi-Omics Data Imbalanced Class Handling and Classification

}
\author{
    Ibrahim Al-Hurani \\
    Affiliation \\
    Lakehead University\\
    Department of Electrical and Computer Engineering\\
    Thunder Bay, ON P7B 5E1, Canada \\
    \texttt{ialhura@lakeheadu.ca} \\
    \And
    Abedalrhman Alkhateeb \\
    Affiliation \\
    Lakehead University \\
    Department of Computer Science\\
    Thunder Bay, ON P7B 5E1, Canada\\
    \texttt{aalkhate@lakeheadu.ca} \\
    \And
    Salama Ikki \\
    Affiliation \\
    Lakehead University\\
    Department of Electrical and Computer Engineering\\
    Thunder Bay, ON P7B 5E1, Canada \\
   \texttt{sikki@lakeheadu.ca} \\
}
\begin{document}
\newacronym{ngs}{NGS}{Next Generation Sequencing }
\newacronym {dr}{DR}{Dimensionality Reduction}
\newacronym{pca}{PCA}{principle component analysis}
\newacronym{svm}{SVM}{Support Vector Machine}
\newacronym{pacmap}{PACMAP}{Probabilistic Approximation model with Controlled Mapping}
\newacronym{smote}{SMOTE}{Synthetic Minority Oversampling TEchnique }
\newacronym{adasyn}{ADASYN}{ADAptive SYNthetic sampling}
\newacronym{gan}{GAN}{Generative Adversarial Networks}
\newacronym{fdr}{FDR}{False Discovery Rates}
\newacronym{hugo}{HUGO}{Human Genome Organization}
\newacronym{deg}{DEG}{differentially expressed genes}
\newacronym{relu}{ReLU}{rectified linear unit}
\newacronym{mse}{MSE}{mean squared error}
\newacronym{tcga}{TCGA}{Cancer Genome Atlas}
\newacronym{brca}{BRCA}{Breast Invasive Carcinoma}
\newacronym{blca}{BLCA}{Bladder Urothelial Carcinoma}
\newacronym{cna}{CNA}{Copy Number Alteration}
\newacronym{ann}{ANN}{Artificial Neural Network}
\newacronym{nmf}{NMF}{Non-negative Matrix Factorization}
\newacronym{ga}{GA}{Genetic Algorithms}
\maketitle
\begin{abstract}

In the relentless efforts in enhancing medical diagnostics, the integration of state-of-the-art machine learning methodologies has emerged as a promising research area. In molecular biology, there has been an explosion of data generated from multi-omics sequencing. The advent sequencing equipment can provide large number of complicated measurements per one experiment. Therefore, traditional statistical methods face challenging tasks when dealing with such high dimensional data. However, most of the information contained in these datasets is redundant or unrelated and can be effectively reduced to significantly fewer variables without losing much information. Dimensionality reduction techniques are mathematical procedures that allow for this reduction; they have largely been developed through statistics and machine learning disciplines. The other challenge in medical datasets is having an imbalanced number of samples in the classes, which leads to biased results in machine learning models. 
This study, focused on tackling these challenges in a neural network that incorporates autoencoder to extract latent space of the features, and \gls{gan} to generate synthetic samples. Latent space is the reduced dimensional space that captures the meaningful features of the original data. Our model starts with feature selection to select the discriminative features before feeding them to the neural network. Then, the model predicts the outcome of cancer for different datasets. The proposed model outperformed other existing models by scoring accuracy of 95.09\% for bladder cancer dataset and 88.82\% for the breast cancer dataset.
\end{abstract}
keywords can be removed
\keywords{Autoenconder, generative adversarial networks, multi-omics data integration, cancer, dimensionality reduction, class imbalance, classification.}

\section{Introduction}
Recent technological advances in \gls{ngs} yield to generate various types of omics data, and omics platforms are becoming increasingly affordable. 
The continuous progress in \gls{ngs} technologies has revolutionized genomics, enabling the generation of diverse omics data at an unprecedented scale \cite{wang2019toward}. Overall, the convergence of advances in \gls{ngs}, enabled comprehensive investigations into complex biological systems.
\gls{dr} techniques have been used in integrating multi-omics data to predict the cancer outcome \cite{alghanim2023machine,qattous2024pacmap}. Alghanim et al. introduced (SMOTE-SVM-RBF) DR embedding model to integrate multi-omics data to predict menopausal status in breast cancer. The model applied \gls{pca} to extract the features latent space of the omics, then integrates them into \gls{svm} \cite{alghanim2023machine}. Qattous et al. proposed the \gls{pacmap} as an embedding layer for multi-omics data integration, \gls{pacmap} converts each omic into one color in a 2D RGB representation. As a result of the RGB limitation, only three omics can be integrated. \gls{pca} is a linear \gls{dr} technique that struggles to represent complex, non-linear structures present in datasets. \gls{pacmap} is probabilistic framework which aims to preserve both the local and global structures of the data, it involves some randomization in its algorithm, which my lead to generate different results each time it runs.

Autoencoders have demonstrated superiority over traditional DR methods in certain scenarios due to their ability to capture complex, non-linear relationships within data. Al-Ghafer et al. proposed (NMF-GA) model that combines \gls{nmf} \gls{ga} to extract meaningful patterns efficiently from high-dimensional datasets \cite{NMF2024}. The model's performance fell short, achieving an accuracy of approximately 78\%.
An autoencoder is an artificial neural network that learns efficient data coding unsupervised. An autoencoder consists of two parts: the encoder and the decoder. The encoder generates reduced feature representations from initial inputs using hidden layers. The decoder reconstructs the original input from an encoder's output by minimizing the loss function \cite{rumelhart1986learning}. The autoencoder converts high-dimensional data to low-dimensional data. Therefore, the autoencoder is especially useful for noise removal, feature extraction, compression, and similar tasks \cite{xhafa2021machine}.

In general, supervised learning algorithms face a challenge where one class has a large number of samples (majority class) and another class has fewer samples (minority class), which is know as class imbalance. Since prediction algorithms tend to favor the majority class, the results in such cases are questionable in terms of reliability and validity \cite{batista2000applying}. The healthcare sector is one of the domains where class imbalance is more common and has a greater impact in the overall performance of the models. A number of techniques have been developed in the literature for dealing with imbalanced data. Synthetic oversampling methods generate new synthetic instances in order to balance a dataset, including \gls{smote} \cite{chawla2002smote}, \gls{adasyn} \cite{he2008adasyn} and \gls{gan}s \cite{goodfellow2014generative}.
\gls{smote} and \gls{adasyn} are data sampling algorithms based on class distributions. Both begin by identifying the minority class. \gls{smote} introduces class imbalance during the \(\textit{k}-\) nearest neighbor selection process, whereas \gls{adasyn} applies a revised version \(\textit{k}-\) 
 nearest neighbors based on weighting scheme. Alghanim et al. compared both \gls{smote} and \gls{adasyn} to generate synthetic multi-omics data and concluded that SMOTE synthetic samples showed more ties to the original minority class distribution \cite{alghanim2023machine}. Unlike traditional data sampling algorithms, \gls{gan} learns to generate synthetic samples that resemble the training data, while the discriminator learns to distinguish between real and generated samples. Through an iterative process, the generator and discriminator compete with each other, improving over time to generate more realistic samples \cite{goodfellow2014generative}.

In this study, a latent space is extracted from each of the omic datasets and merged into a shared latent space using an autoencoder. Then, to handle imbalance data classification, the minority class is up-sampled using GAN. We use the resulting shared latent space dataset to feed artificial neural networks that predicts cancer outcomes in two different datasets.  

\section{Materials and Methods}
\begin{figure} [!ht]
  \centering
  \fbox{\includegraphics[width=0.8\linewidth]{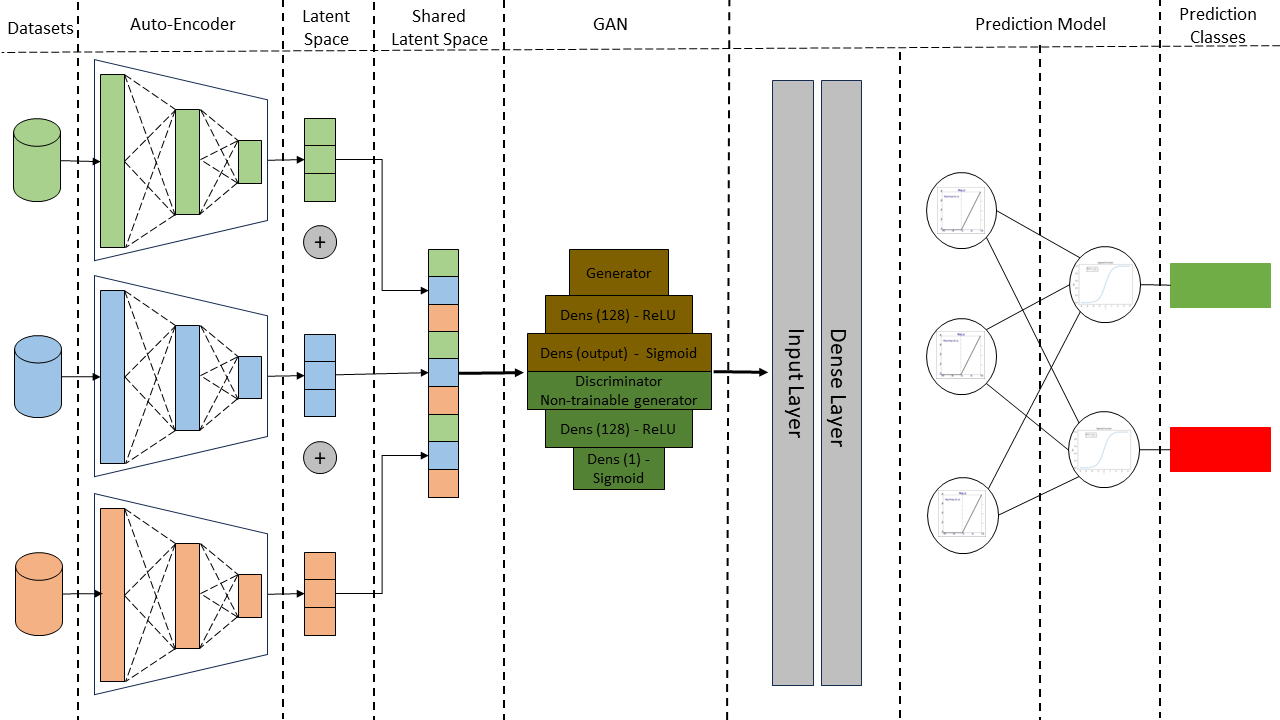}}
  \caption{The workflow of the proposed model.}
  \label{GAN_Figur}
\end{figure}

As a first step, we pre-process the three omics dataset using the autoencoder to extract latent space, and then generate more minor class samples to address the imbalanced classes. 
The proposed model can be seen schematically in Figure \ref{GAN_Figur}.
\subsection{Materials}
The proposed model is applied to two publicly available datasets, the first is the \gls{tcga} \gls{brca} dataset, which contains DNA Methylation, \gls{cna}, and Gene information \cite{ciriello2015comprehensive}. The second is the \gls{tcga} \gls{blca} dataset, which contains DNA Methylation, \gls{cna} and mRNA \cite{gao2013integrative}. For the \gls{brca} dataset, we compared the result of applying the proposed model with the state of art SMOTE-SVM-RBF model \cite{alghanim2023machine}. While for the \gls{blca} dataset, we compared the results with NMF-GA model \cite{NMF2024}. For both datasets the three omics were downloaded from Cbioportal \cite{cerami2012cbio}. 
\subsection{Pre-processing}
The \gls{brca} dataset was filtered by removing gene expression features that had less than 0.2\% variability, which reduced the number from around 39,000 to around 16,000 features. After normalizing all three omics datasets using z-score normalization, genes that did not adhere to the \gls{hugo} format were removed. To identify significant mutations in breast cancer genes, the MutsigCV algorithm was used. Using this algorithm, \gls{fdr}s are calculated for genes with p-values < 0.05. Genes with \gls{fdr}s < 0.1 were considered significantly mutated, resulting in the selection of 14 mutated genes from the MutsigCV output for inclusion in our study.
The \gls{blca} dataset features are all normalized using z-score. In order to select those features with significant differences in means between the two classes, we perform a t-test with p<0.05 (CI: 95\%) for all features in the three omics. After selecting the \gls{deg}s, the fold-change test was performed on the \gls{deg}s, which is a statistical method that measures the extent of variation in a variable’s value between two conditions or treatments. The logarithm of base two is then taken to the ratio to select the feature that is either doubled or halved between the two classes. Benjamini \& Hochberg test \cite{benjamini1995controlling} was used to adjusts the p-value in multiple hypothesis testing to control the \gls{fdr}. In this study, the selected FDR value is 0.01, which filters out the features with an adjusted p-value greater than that threshold.
\subsection{Performance Measurements}
To evaluate the classification models, we used the following performance metrics: 
\begin{equation}
{ Accuracy }=\frac{TN + TP}{TN + TP + FP + FN}
\end{equation}
\begin{equation}
{ Recall}=\frac{TP}{TP + FN}
\end{equation}
\begin{equation}
{ ROC }=\frac{P(x | positive)}{P(x | negative)}
\end{equation}
\begin{equation}
{ Precision }=\frac{TP}{TP + FP}
\end{equation}
\begin{equation}
{ F1-Score } = 2 * \frac{{ Precision } * { Recall }}{{ Precision }+ { Recall }}
\end{equation}
\subsection{Methods}
\subsubsection{Autoencoder}

The encoder, implemented as a dense neural network layer with \gls{relu} activation, compresses the input data into a lower-dimensional representation, this layer reduces the dimensionality of the input while capturing its latent space. The mathematical representation of the encoder is shown in equation \ref{eq:Encoder}:
\begin{equation}
\text{Encoder}:\mathbf{h} = f_{\text{encoder}}(\mathbf{x}) = \phi(\mathbf{W}_1\mathbf{x} + \mathbf{b}_1)
\label{eq:Encoder}
\end{equation}
Where:
\begin{itemize}
    \item $\mathbf{x}$ is the input data,
    \item $\mathbf{h}$ is the encoded representation (latent space),
    \item $f_{\text{encoder}}$ is the encoder function,
    \item $\mathbf{W}_1$ and $\mathbf{b}_1$ are the weights and biases of the encoder layer, respectively,
    \item $\phi$ is a ReLU activation function, which returns the maximum between and 0. 
\end{itemize}
The decoder, a dense neural network layer with Sigmoid activation, reconstructs the original input from the encoded representation. The mathematical representation of the decoder is show in equation \ref{eq:Decoder}:
\begin{equation}
\text{Decoder}: \mathbf{x'} = f_{\text{decoder}}(\mathbf{h}) = \phi(\mathbf{W}_2\mathbf{h} + \mathbf{b}_2)
\label{eq:Decoder}
\end{equation}
Where:
\begin{itemize}
    \item $\mathbf{x'}$ is the reconstructed output,
    \item $f_{\text{decoder}}$ is the decoder function,
    \item $\mathbf{W}_2$ and $\mathbf{b}_2$ are the weights and biases of the decoder layer, respectively.
\end{itemize}
Throughout training, the autoencoders minimize the \gls{mse} between the input and the output, ensuring accurate reconstruction. After training, the autoencoders are used to generate latent space representations for the respective input data types. During training, the parameters $(\mathbf{W}_1, \mathbf{b}_1, \mathbf{W}_2, \mathbf{b}_2)$ are optimized to minimize the \gls{mse} loss between the input $(\mathbf{x})$ and the output $(\mathbf{x'})$. \gls{mse} be calculated as show in equation \ref{eq:MSE}:
\begin{equation}
\mathcal{L}(\mathbf{x}, \mathbf{x'}) = \frac{1}{N}\sum_{i=1}^{N}(\mathbf{x}_i - \mathbf{x'}_i)^2
\label{eq:MSE}
\end{equation}
Where:
\begin{itemize}
\item $N$ is the number of data points in the batch.
\end{itemize}
A separate autoencoder is applied to each omic, and then the latent spaces generated from all of them are merged into one shared latent space.

\subsubsection{Generative Adversarial Networks (GAN)}
In this step, \gls{gan} is utilized to up-sample the minor class from the shared latent space samples. The generator $G$ maps random noise $z$ sampled from a prior distribution $p_z(z)$ to synthetic samples $\hat{x}$:
\[ \hat{x} = G(z) \]

The discriminator $D$ takes a sample $x$ and outputs a probability $D(x)$ indicating the likelihood that $x$ is a real sample:
\[ D(x) \]
The generator network is constructed by first initializing a sequential model, that consists of:
\begin{itemize}
    \item Dense layer with 128 neurons.
    \item A \gls{relu} activation function, which helps the model learn complex patterns in the data by allowing it to model non-linear relationships.
    \item Dense layer with a specific output dimension and a Sigmoid activation function is added to produce the final output, which generates synthetic samples.
\end{itemize}
The discriminator network is constructed using a sequential model, that consists of:
\begin{itemize}
\item Dense layer with 128 neurons and \gls{relu} activation function to process the input data.
\item Dense layer with a single neuron and Sigmoid activation function is added to classify the input samples as real or fake. 
\end{itemize}

These networks are key components of the \gls{gan} framework, where the generator creates synthetic samples and the discriminator tries to differentiate between real and fake data. By fostering a competitive learning process, \gls{gan} boosts the generator's ability to produce alike realistic data.

The generator network is defined as:
 \begin{equation}
    G(z) = \sigma(\phi(W_g z + b_g))
    \label{eq:Generator_Network}
    \end{equation}
Where:
\begin{itemize}
    \item \( G(z) \) represents the synthetic sample generated by the generator.
    \item \( z \) is the input noise vector,
    \item \( W_g \) and \( b_g \) are the weights and biases of the generator's dense layers,
    \item ReLU (\( \phi \)) is the Rectified Linear Unit activation function, which is defined as \( \phi(x) = \max(0, x) \), helping the model learn non-linear patterns,
    \item Sigmoid (\( \sigma \)) is the Sigmoid activation function, which squashes the output values between 0 and 1, suitable for generating data samples.
\end{itemize}
For the discriminator network:
 \begin{equation}
    D(x) = \sigma(\phi(W_d x + b_d))
    \label{eq:Discriminator_Network}
    \end{equation}

The objective of the \gls{gan} is to train the generator to produce samples that are indistinguishable from real samples, and to train the discriminator to distinguish between real and fake samples. This adversarial training process can be formulated as a minimax game:
\begin{equation}
\min_G \max_D V(D, G) = \mathbb{E}_{x \sim p_{\text{data}}(x)} [\log D(x)] + \mathbb{E}_{z \sim p_z(z)} [\log(1 - D(G(z)))]
\label{eq:MinMax}
\end{equation}

Where:
\begin{itemize}
    \item $\mathbb{E}$ denotes the expectation over real data $x$ and noise $z$,
    \item $p_{\text{data}}(x)$ is the distribution of real data,
    \item $p_z(z)$ is the prior distribution of noise.
\end{itemize}

\textit{Training Process}

During training, the discriminator and generator are alternatively updated as the following:
\begin{itemize}
    \item Discriminator Training: The discriminator is trained to maximize its ability to differentiate between real and fake samples. It is trained on both real samples $x$ and generated samples $\hat{x}$, and its loss is calculated using binary cross-entropy:
    \begin{equation}
    \mathcal{L}_{\text{D}} = -\left(\log D(x) + \log(1 - D(\hat{x}))\right)
    \label{eq:Discriminator_Training}
    \end{equation}
    \item Generator Training: The generator is trained to minimize the discriminator's ability to distinguish between real and fake samples. Its loss is calculated using the discriminator's output when fed with generated samples:
    \begin{equation}
    \mathcal{L}_{\text{G}} = -\log(D(G(z)))
    \label{eq:Generator_Training}
    \end{equation}
\end{itemize}

\subsubsection{Prediction Model}
Firstly, we split the data into training and testing sets based on a split ratio of 0.8, that is used to construct the prediction model. A  network is built of two dense layers. The first is with 128 neurons and ReLU activation function, and the other with 1 neuron and Sigmoid activation function. The model utilizes Adam optimizer\cite{kingma2014adam} and binary cross-entropy loss function to optimize the accuracy. The model is trained with parameters specifying the number of training epochs of 10, batch size of 32, and a validation split of 0.2. The Sigmoid function returns the probability of the class which represents the predicted class.  

\section{Results}
 The performance measurements for the BRCA dataset are summarized in Table \ref{tab:Compare_Proposed_to_SMOT-SVM-RBF_on_BRCA}. Our proposed model exhibits exceptional performance across various metrics compared to the SMOTE-SVM-RBF classifier. Notably, the proposed model achieves an average accuracy of 95.09\%, showcasing a significant improvement over the 78.03\% accuracy achieved by the SMOTE-SVM-RBF classifier. Furthermore, the proposed model demonstrates a perfect average AUC of 1.00, indicating excellent discriminatory power, while the SMOTE-SVM-RBF classifier achieves a considerably lower AUC of 0.7805. In terms of precision, the proposed model achieves a perfect precision of 100\%, indicating its ability to accurately identify positive instances, whereas the SMOTE-SVM-RBF classifier achieves a precision of 77.58\%. Additionally, our model achieves a recall of 81.48\%, suggesting its effectiveness in capturing true positive instances, compared to a recall of 80.24\% in the SMOTE-SVM-RBF classifier. Finally, the proposed model attains an average F1-score of 89.79\%, surpassing the 78.84\% F1-score achieved by the SMOTE-SVM-RBF classifier. 

\begin{table} [htbp]
 \caption{The performance measurements for the BRCA dataset} 
  \centering
  \begin{tabular}{llllll}
    \toprule
    Classifier & Average accuracy & Average AUC ($\mu$m) &  Precision &  Recall & Average F1-score \\
    \midrule
    Proposed Model & 0.95098 & 1.0000 & 1.0000 & 0.81481 & 0.89796     \\
    SMOT-SVM-RBF & 0.78036 & 0.78050 & 0.7758 & 0.8024  & 0.78840    \\
        \bottomrule
  \end{tabular}
  \label{tab:Compare_Proposed_to_SMOT-SVM-RBF_on_BRCA}
\end{table} 


The performance measurements for the BLCA dataset are summarized in Table \ref{tab:Compare_Proposed_to_NMF-GA_on_BRCA}. Our proposed model exhibits promising performance across various metrics compared to the NMF-GA classifier. Specifically, the proposed model achieves an average accuracy of 88.82\%, significantly outperforming the NMF-GA classifier, which obtains an accuracy of 78.03\%. Additionally, the proposed model demonstrates a higher average AUC of 91.21\% compared to 78.05\% in the NMF-GA classifier. In terms of precision, the proposed model achieves 85.41\%, indicating its ability to accurately identify positive instances, while the NMF-GA classifier achieves a slightly lower precision of 77.58\%. Moreover, our model achieves a recall of 75.92\%, suggesting its effectiveness in capturing true positive instances, whereas the NMF-GA classifier achieves a higher recall of 80.24\%. Finally, the proposed model attains an average F1-score of 80.39\%, surpassing the 78.84\% F1-score achieved by the NMF-GA classifier.
 
\begin{table} [htbp]
 \caption{The performance measurements for the BLCA dataset} 
  \centering
  \begin{tabular}{llllll}
    \toprule
    Classifier & Average accuracy & Average AUC ($\mu$m) &  Precision &  Recall & Average F1-score \\
    \midrule
    Proposed Model & 0.88827 & 0.91211 & 0.85417 & 0.75926  & 0.80392\\ 
    NMF-GA    & 0.78036 & 0.78050 & 0.7758 & 0.8024  & 0.78840    \\
        \bottomrule
  \end{tabular}
  \label{tab:Compare_Proposed_to_NMF-GA_on_BRCA}
\end{table} 



\begin{figure} [htbp]
  \centering
  \includegraphics[width=0.8\linewidth]{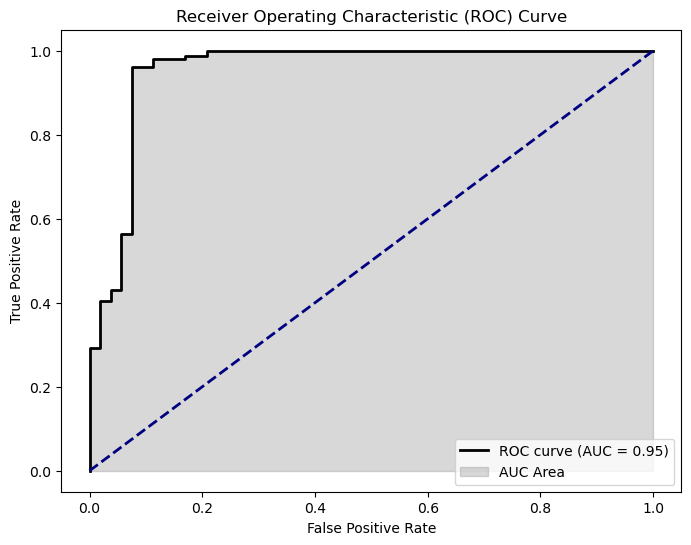}
  \caption{The AUCROC Curve for running the proposed model on the BRCA dataset.}
  \label{Fig:GAN-ENC-Breast-Receiver- Operating-Characteristic-ROC-Curve}
\end{figure}
\begin{figure} [htbp]
  \centering
  \includegraphics[width=0.8\linewidth]{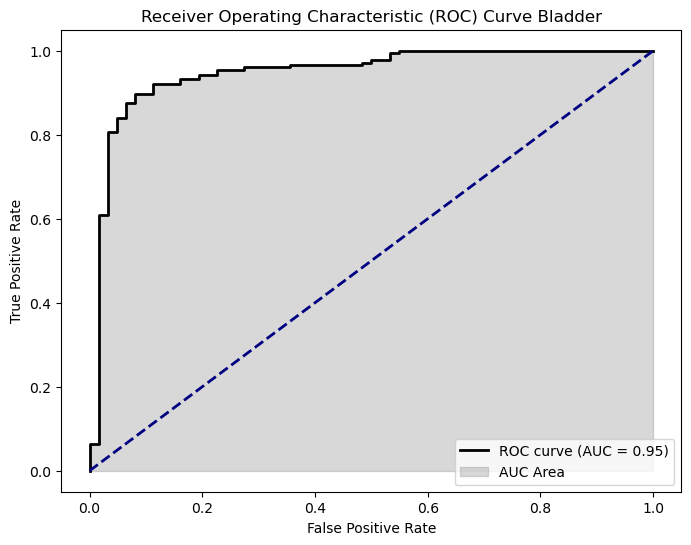}
  \caption{The AUCROC Curve for running the proposed model on the BLCA dataset.}
  \label{Fig:GAN-ENC-Bladder-ROC-Curve}
\end{figure}

Figures \ref{Fig:GAN-ENC-Breast-Receiver- Operating-Characteristic-ROC-Curve}-\ref{Fig:GAN-ENC-Bladder-ROC-Curve} show the area under the curve for the receiver operating characteristic (AUCROC curve) for running the proposed model on both BRCA and BLCA datasets. AUCROC shows the trade off between the sensitivity (or TPR) versus the specificity (or FPR) across multiple running points.

\section{Conclusion}
This study demonstrates significant advancements in predictive accuracy for cancer outcome prediction. The approach begins with feature selection to extract discriminative features. This is followed by dimensionality reduction using autoencoders to generate a compact latent space representation of the data. Incorporating the GAN framework, the model augments the dataset by generating synthetic samples for the minority class, thus mitigating class imbalance. In the final step, the approach classifies multi-omics data better for cancer prediction.

Experimental results on publicly available datasets demonstrate the effectiveness of our proposed approach. We have outperformed existing models in terms of accuracy, precision, recall, and F1-score, showcasing the robustness and generalizability of our model across different cancer types and omics data.

Overall, our study shows how autoencoders and GANs can enhance machine learning models' predictive capabilities for medical diagnosis. In addition to addressing the challenges posed by imbalanced class distributions and high-dimensional omics data, our approach contributes to precision medicine and personalized healthcare.


\bibliographystyle{plain}
\bibliography{references}


\end{document}